\documentclass[11pt]{article}
\usepackage{nodalida2021}
\usepackage{times}
\usepackage{url}
\usepackage{latexsym}

\usepackage{microtype}

\aclfinalcopy 

\usepackage{graphicx}
\usepackage{tabularx}
\usepackage{soul}
\usepackage{xspace} 
\usepackage{epstopdf}

\usepackage[T1]{fontenc}
\usepackage[utf8]{inputenc}

\usepackage{hyperref}
\usepackage{xstring}
\usepackage{cleveref}

\usepackage{color, colortbl}
\definecolor{Gray}{gray}{0.9}

\newcommand{\ft}{Folketinget\xspace}

\title{The Danish Gigaword Corpus}

\author{Leon Derczynski \\
  ITU Copenhagen \\
  Denmark\\
  \texttt{ld@itu.dk} \\\And
  Manuel R. Ciosici \\
  USC Information Sciences Institute \\
  USA\\
  \texttt{manuelc@isi.edu}\AND
  Rebekah Baglini\\
  Aarhus University\\
  Denmark\\\And
  Morten H. Christiansen\\
  Aarhus University \& Cornell University\\
  Denmark\\\AND
  Jacob Aarup Dalsgaard\\
  Aarhus University\\
  Denmark\\\And
  Riccardo Fusaroli\\
  Aarhus University\\
  Denmark\\\AND
  Peter Juel Henrichsen\\
  Danish Language Council\\
  Denmark\\\And
  Rasmus Hvingelby\\
  Alexandra Institute\\
  Denmark\\\AND
  Andreas Kirkedal\\
  ITU Copenhagen\\
  Denmark\\\And
  Alex Speed Kjeldsen\\
  University of Copenhagen\\
  Denmark\\\AND
  Claus Ladefoged\\
  TV2 Regionerne\\
  Denmark\\\And
  Finn Årup Nielsen\\
  Technical University of Denmark\\
  Denmark\\\AND
  Jens Madsen\\
  Karnov Group\\
  Denmark\\\And
  Malte Lau Petersen\\
  Aarhus University\\
  Denmark\\\AND
  Jonathan Hvithamar Rystrøm\\
  Aarhus University\\
  Denmark\\\And
  Daniel Varab\\
  Novo Nordisk \& ITU Copenhagen\\
  Denmark}

\date{}

\begin{document}
\maketitle
\begin{abstract}
 Danish language technology has been hindered by a lack of broad-coverage corpora at the scale modern NLP prefers. This paper describes the Danish Gigaword Corpus, the result of a focused effort to provide a diverse and freely-available one billion word corpus of Danish text. The Danish Gigaword corpus covers a wide array of time periods, domains, speakers' socio-economic status, and Danish dialects.
\end{abstract}

\section{Introduction}

It is hard to develop good general-purpose language processing tools without a corpus that is broadly representative of the target language. Further, developing high-performance deep learning models requires hundreds of millions of tokens~\cite{radfordlanguage,JMLR:v21:20-074}. 
To address this gap for Danish, a North Germanic/Scandinavian language spoken primarily in Denmark, we propose an open giga-word corpus. This corpus is free to download and use, thus enabling researchers and organizations to further develop Danish NLP without worrying about licensing fees. The corpus is a first necessary step to allow Danish speakers to receive the many benefits of the powerful range of NLP technologies.

This paper details the Danish Gigaword Corpus~({\sc Dagw}), a billion-word corpus of language across various dimensions, including modality, time, setting, and place.

It is tricky to collect such a corpus automatically: automatic language identification tools confound closely related languages, especially Danish and Bokm{\aa}l, and are likely to miss important data~\cite{radfordlanguage,nordicdsl}. Existing representations underperform for Danish: the multilingual FastText embeddings~\cite{joulin2018loss} miss core Danish words such as ``tr{\ae}ls'';
Multilingual BERT lacks sufficient support for the Danish vowel ``{\aa}''.\footnote{BotXO maintains a Danish BERT instance at {\tt https://github.com/botxo/nordic\_bert}. This model was trained exclusively on uncurated web text and, therefore, (a) has a spurious understanding of Danish among other languages and (b) is particularly susceptible to the kind of toxic language identified by \citet{gehman-etal-2020-realtoxicityprompts}.}

To remedy this situation, we propose a Danish Gigaword Corpus. The overriding goals are to create a dataset that is (1) representative, (2) accessible, and (3) a general-purpose corpus for Danish.

\section{Background}

Today's NLP is generally data-intensive, meaning that large representative corpora tend to correlate with better models and better processing results.
However, large representative corpora are available for only a small set of languages; there are fewer than ten manually-compiled gigaword-scale corpora, for example, and none for Danish.

Several substantial Danish text corpora have been compiled during recent decades. CLARIN-DK offers a variety of individual corpora of varying genres, annotations, and writing times. However, non-commercial licensing restricts corpus usage.
Some major Danish corpora are related to dictionary production, as is the case for the 56 million words Korpus-DK available for search at the dictionary site ordnet.dk.\footnote{\url{http://ordnet.dk}}
Leipzig Corpora Collection assembles Danish corpora from the Web, news sites, and Wikipedia~\cite{goldhahn-etal-2012-building}. The combined size of these corpora is orders of magnitude smaller than The Danish Gigaword Corpus. By themselves, these corpora do not meet the data size needs of modern language models.

Modern language models like T5~\cite{JMLR:v21:20-074} and GPT2~\cite{radfordlanguage} are text-hungry, making automatic corpora construction attractive. Massive, monolithic, automatically collected datasets of web content, such as Common Crawl, support the training of large language models but suffer from quality issues~\cite{radfordlanguage} and bias~\cite{Ferrer2020}. Models trained exclusively with such data quickly delve into generating toxic language~\cite{gehman-etal-2020-realtoxicityprompts}. Furthermore, the Danish section of Common Crawl is plagued by significant amounts of non-Danish content, in part due to the pervasive confusion between Danish and Norwegian Bokm{\aa}l by highly multilingual language ID classifiers~\cite{nordicdsl}. Datasets derived exclusively from Common Crawl also have a bias toward webspeak and content from recent years, leaving models built over them sub-optimally prepared to process older Danish.

The lack of a large and qualitative Danish corpus causes Danish NLP tools to lag behind equivalent tools for better-resourced languages, and the gap is increasing~\cite{0e564c1503f64fef910d72e1106d784d,kirkedal2019lacunae,kirchmeier-etal-2020-world}.

The first gigaword corpus was the English Gigaword~\cite{graff2003english}, consisting of roughly one billion ($10^9$) words of English-language newswire text. The content was single-genre, national and global newswire, published between 1994 and 2002. Other gigaword corpora emerged later, for French, Arabic, Chinese, and Spanish. Even Icelandic, a language with just over $360\,000$ speakers, has a healthy gigaword project~\cite{steingrimsson2018risamalheild}.

\section{Linguistic diversity}

\begin{figure}
    \centering
    \includegraphics[width=0.5\textwidth]{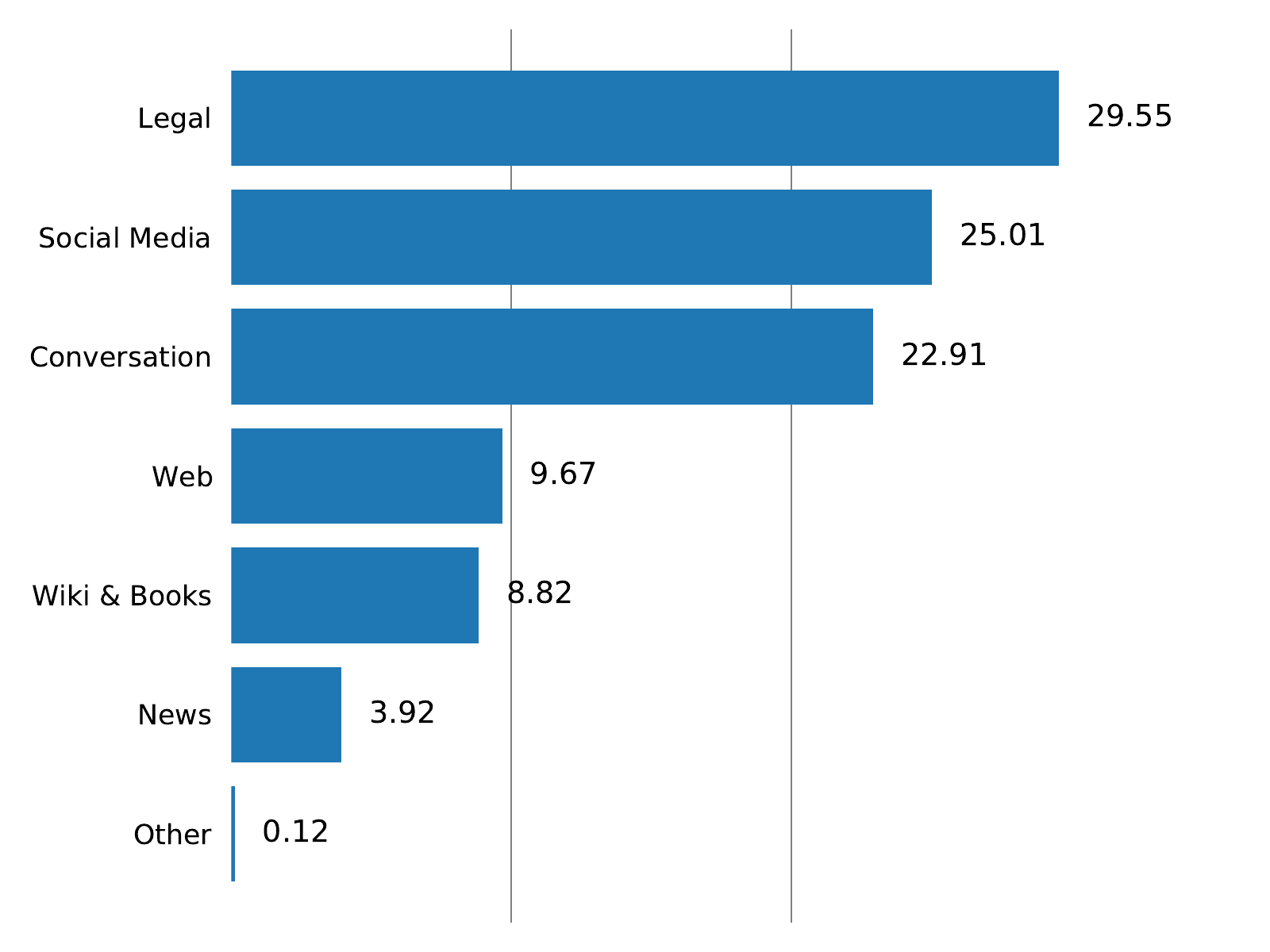}
    \caption{Content by domain (\% of corpus).}
    \label{fig:domains}
\end{figure}

For a corpus to be useful for a wide range of applications, it must include a wide range of language, mixing domains, speakers, and styles~\cite{biber1993representativeness}. Failing to do this can lead to severe deficiencies in the data. For example, when NLP work started on social media text, the Wall Street Journal-trained part of speech taggers missed essential words such as ``Internet'' (due to the articles being from the late eighties and early nineties) and ``bake'', due to their domain.

Common Crawl's undirected collection of content often over-represents some dialects at the expense of other dialects.  GeoWAC~\cite{dunn-adams:2020:LREC} uses demographic information to construct English corpora that balance dialects. Unfortunately, a demographic- and Web-based approach underrepresents Danish dialects such as the endangered Bornholmsk dialect~\cite{mortensenbornholmske}, which is almost absent from the Web.

These deficiencies do not form a solid basis for general-purpose NLP. So the Danish Gigaword Corpus captures and distributes as broad a range of Danish language use as possible, explicitly including language from a variety of settings (long-form writing, novels, social media, speeches, spontaneous speech), domains~(news, politics, fiction, health, social media, law, finance), time periods~(from the 1700s to present day), registers~(formal, informal), and dialects~(including, e.g., Bornholmsk and S{\o}nderjysk).

\begin{table*}[t]
    \footnotesize
    \centering
    \begin{tabular}{l|l|l|l|l|l|r}
        & {\bf Date}  & {\bf Form} & {\bf Domain}  & {\bf Dialect} & {\bf Socioeconomic status} & {\bf Size (M)} \\  
        \hline

        \multicolumn{7}{c}{} \\
        \rowcolor{Gray} \multicolumn{6}{c}{{\bf Legal}} & {\bf 308.8} \\
        Retsinformation  & contemporary & written   & Laws & legal & high & 188.4 \\
        Skat.dk  & contemporary & written   & Tax code & legal & high & 52.8 \\
        H-Sø     &  contemporary  & written  & Court cases & mixed & mixed & 67.6 \\
        \hline

        \multicolumn{7}{c}{} \\ 
        \rowcolor{Gray} \multicolumn{6}{c}{{\bf Social Media}} & {\bf 261.4} \\
        Hestenettet   	    &  contemporary  & written  & forum & mixed & mixed & 228.9 \\
        General Discussions  &  2\,019 - 2\,020  & written  & Twitter & mixed & mixed & 32.0 \\
        Parliament Elections  &  2\,019  & written  & Twitter & mixed & mixed & 0.5 \\
        \hline

        \multicolumn{7}{c}{} \\
        \rowcolor{Gray} \multicolumn{6}{c}{{\bf Conversation}} & {\bf 239.4} \\
        OpenSubtitles   	& contemporary       & spoken   & Movie subtitles & mixed & mixed & 130.1  \\
        \ft            & 2\,009 - 2\,019    & spoken   & Debates & rigsdansk & high & 60.6 \\
        Europarl   	& 2\,004 - 2\,008    & spoken   & Debates & standard & mixed & 47.8 \\
        Spontaneous speech  & 2\,019   & spoken   & Conversation & mixed & mixed & 0.7 \\
        NAAT   	    &  1930 - now  & spoken  & Speeches & rigsdansk & high & 0.2 \\
        \hline

        \multicolumn{7}{c}{} \\
        \rowcolor{Gray} \multicolumn{6}{c}{{\bf Web}} & {\bf 101.0} \\
        Common Crawl  &  contemporary  & written  & Web & mixed & mixed & 101.0 \\
        \hline

        \multicolumn{7}{c}{} \\
        \rowcolor{Gray} \multicolumn{6}{c}{{\bf Wiki \& Books}} & {\bf 92.2} \\
        Wikipedia  & 2\,019 - 2\,020    & written  & Encyclopaedic & standard & mixed & 55.6  \\
        Danish Literature  & 1\,700 - now    & written  & Literature & standard & mixed & 25.6  \\
        Gutenberg  & 1\,700 - now    & written  & Literature & standard & mixed & 3.2  \\
        WikiBooks  & 2\,019 - 2\,020  & written  & Manuals & standard & mixed & 2.6  \\
        WikiSource  & 1\,700 - now    & written  & Literature & standard & mixed & 2.5  \\
        Johannes V. Jensen   	& -  & written  & JVJ's works & rigsdansk & unknown & 2.1 \\
        Religious texts   	& -  & written  & Religious & rigsdansk & unknown & 0.6 \\
        \hline

        \multicolumn{7}{c}{} \\ 
        \rowcolor{Gray} \multicolumn{6}{c}{{\bf News}} & {\bf 40.0} \\
        TV2R   	    &  2\,015 - 2\,019  & written  & News & rigsdansk & high & 10.0 \\
        DanAvis    &  1\,999 - 2\,003  & written  & News & rigsdansk & medium & 30.0 \\
        \hline

        \hline
        \multicolumn{7}{c}{} \\ 
        \rowcolor{Gray} \multicolumn{6}{c}{{\bf Other}} & {\bf 1.2} \\
        Dasem data\footnote{https://github.com/fnielsen/dasem}   	    & contemporary   & written  & Other & mixed & mixed & 0.7 \\
        Botxt   	    & contemporary   & written  & Other & Bornholmsk & mixed & 0.4 \\
        DDT   	    & contemporary   & written  & Other & mixed & mixed & 0.1 \\
        Sønderjysk   	    & contemporary   & written  & Sønderjysk & Sønderjysk & mixed & 0.02 \\
        \hline
        
        \hline
        \multicolumn{7}{c}{} \\
        \rowcolor{Gray} \multicolumn{6}{c}{{\bf TOTAL}} & {\bf 1\,045} \\
    \end{tabular}
    \caption{Text dimensions by text source in the Danish Gigaword corpus. Size in millions of words.}
    \label{tbl:text_dimensions}
\end{table*}

\section{Dataset construction}

The Danish Gigaword Corpus consists of sections, with each section corresponding to a single source of text. Following prior efforts to construct broad-coverage datasets~\cite{derczynski2016broad}, sections are selected based on how well they help the corpus' coverage of Danish language use over a variety of dimensions, including:
time of authorship;
speech situation;
modality;
domain;
register;
age of utterer;
dialect of utterer;
socio-economic status of utterer.
This is a strong, intentional departure from editions of English Gigaword that focused on newswire. Achieving some degree of representativeness~\cite{biber1993representativeness} requires the inclusion of sources beyond newswire text.
We provide an overview of The Danish Gigaword Corpus's content in \Cref{fig:domains} and detail the sections in \Cref{tbl:text_dimensions} and the appendix.

The Danish Gigaword Corpus follows the definition of genre used by~\newcite{biber1993representativeness}, grounded in ``situationally defined categories'', such as a language style recognized by (or used to define) a community, such as news articles, personal letters, or online chat; a domain as a particular topical focus (or set of foci) that are discussed, such as biomedicine, politics, or gaming; and a medium as the means by which communication is conducted, such as writing, online chat, conversation, and so on. There is a natural overlap between medium and speech situations, but the delineation is beyond this work's scope.

While the goal of {\sc Dagw} is to cover a range of genres, domains, and media, it is difficult to measure the prevalence of each of these across all Danish users, let alone then gather and redistribute this data. Therefore, the goal is to cover something of everything that can be feasibly included, without letting any particularly monolithic combination dominate (in contrast to, e.g., the 100\% written newswire content of English Gigaword v1 or the 100\% Common Crawl content of GeoWAC). Not every intersection between genres, domains, and media can be covered, nor represented proportionally, in the first version of this corpus. 
\Cref{tbl:text_dimensions}  contains an overview of the genres, domains, and modalities included in the Danish Gigaword Corpus.

\subsection{Data and metadata unification}
Each section is contained in one directory, named after the ``prefix'' for the section. Each file in a section represents a single UTF encoded document. Each section contains at least two functional files: one describing how the section is licensed and one describing metadata about each document. For multi-speaker corpus sections, an optional file can contain a dictionary keyed by speaker ID. This assumes speaker IDs are used consistently through all documents in that section. 
\Cref{sec:app:file_format} contains a complete description of the file format.

Sections are managed individually as part of a larger repository of the whole Danish Gigaword Corpus. A validation script helps make sure that the sections comply with the file format.

\subsection{Data protection}

The corpus does not contain ``sensitive'' data as per the GDPR definition; that means no information identifying sexual orientation, political beliefs, religion, or health connected with utterer ID. This is achieved by stripping utterer information from social media content. Thus, data discussing potentially personally sensitive topics, for example, social media around political discussions, is disconnected from personally-identifying information.
Further, social media content is supplied not as plain text but as IDs and code for rehydration, a process where the content is re-downloaded, thus avoiding redistribution of this content and affording social media users the ability to delete their content without it being preserved by Danish Gigaword.

\subsection{Test/Train partitions}

Following the result that fixed test/train splits lead to unreliable results~\cite{gorman-bedrick-2019-need}, we avoid setting explicit test/train partitions in Danish Gigaword. We encourage users to select multiple random test splits. Since the Danish Gigaword is highly diverse, selecting multiple random splits will result in test sets with different biases following best practices~\cite{Sogaard2020}.

\subsection{Licensing}
\label{sec:licensing}

All corpus parts are licensed openly, for free distribution. We implement this with a mixture of Creative Commons general license (CC0) and CC-BY. 
Some older corpora (e.g., \newcite{kromann2003danish}) used the right under Danish copyright law to cite small excerpts of up to 250 words from published articles. While this is a creative solution to sharing digital language data, Danish Gigaword uses almost exclusively whole articles, as they are easier to work with, providing full context.

\section{Distribution and sustainability}

As mentioned earlier in this paper and by \citet{kirkedal2019lacunae,kirchmeier19,kirchmeier-etal-2020-world}, one problem that plagues Danish NLP is a lack of large accessible corpora. 
To address this and maintain strict licensing standards that permit open and free redistribution, Danish Gigaword Corpus is hosted and freely distributed via \url{https://gigaword.dk/}. Alternative downloads will be provided through major dataset distribution services at each significant release.

{\sc Dagw} is an intrinsically open project. In a bid to improve and uphold its relevance at a broad level, the current group of participants covers academia, industry, and the public sector. However, the {\sc Dagw} project is also volunteer-led and volunteer-driven, which brings intrinsic risk. Aside from cross-sector involvement, the {\sc Dagw} project attempts to mitigate that risk through licensing, distribution, membership, community, and data integrity policies.

Strategically, the corpus strives for an improved balance. The contents in the first release, with this paper, reflect the data that is available in Denmark. Data that is legally required to be open and unlicensed dominates the corpus, reflecting the current state of text sharing in Denmark. We hope that this will become less conservative over time and particularly look forward to further donations of newswire and literature, so that NLP for Danish can start to offer Danish speakers improved technology.

The data is licensed CC-BY and CC0, which gives it broad reach and applicability, and makes it easier for stakeholders to join than copyleft or non-commercial licenses, such as GPL or CC-NC, would. It also improves distribution prospects: because of this licensing choice, {\sc Dagw} can be hosted at a third-party research data repository like Zenodo or Figshare, shifting the responsibility for data hosting and provision to specialized third parties. The {\sc Dagw} project also maintains an open policy, with any qualified stakeholders welcome to join, especially if there is a compatible donation of data. Denmark's size helps keep a manageable community. The Danish Gigaword also fosters community involvement by publishing results -- for example, this paper. Finally, a small toolkit is included in the project's Github repository for automatic validation of any committed data, ensuring content integrity, quality, and uniformity.

\section{Conclusion and Future Work}

In Denmark, natural language processing is nascent and growing faster and faster. Content restrictions and conservative licensing abound. This paper presents the Danish Gigaword Corpus, a unified effort across many institutions and many Danish speakers to construct a billion-word corpus representing the language. It aims to be useful to a maximally broad and diverse group of users. 

The Danish Gigaword Corpus is an active project. There is continuing effort to add sources that enhance the corpus' breadth, including fiction, older works from the 1800s, and newswire. {\sc dagw} continues past the first billion words, with data always released under Creative Commons license and freely distributed via \url{https://gigaword.dk/}.

We hope that this concrete and significant contribution benefits anyone working with Danish NLP or performing other linguistic activities and encourages others to publish language resources openly.

\section*{Acknowledgments}
This work was not supported by any funded project or university initiative, but rather was a labour of love by the first two, \textit{``fremmedarbejder''}, \textit{``tosprogede''} authors, who thought Denmark really ought to have a decent-sized open corpus of Danish. And now it has. We are extremely grateful for the generous contributions of time, effort, and data from so many that made this project possible.

\bibliography{dagw}
\bibliographystyle{acl_natbib}

\appendix

\section{Detailed corpus description}
\label{sec:app:corpus_table}

Here we detail some of the sections included in the corpus, specifying what they bring to the dataset to make it a rich resource covering a wide range of lexical, syntactic, and sociolinguistic phenomena expressed by Danish users. \Cref{tbl:text_dimensions} provides an overview of the corpus.

\subsection{TV2 Regionerne}

This section is a contemporary Danish newswire sample: approximately 50\,000 full newswire articles published between 2010 and 2019. It contains articles of regional interest, written following editorial standards. This section's value is in both its temporal variation, covering a decade of events, and its spatial variation, covering many local events across most of Denmark (TV2 Bornholm is excluded). As a result of local event coverage, the section contains many locally relevant named entities, which might otherwise not be present in a dataset of national news.

\subsection{Folketinget}

The Danish parliament (\ft) keeps a record of all meetings in the parliament hall.\footnote{There are no records of committee meetings or \emph{samr\aa d}.} All records have a transcript produced by commercial Automatic Speech Recognition (ASR) followed by post-editing by linguists employed by \ft for intelligibility, i.e., edit out dysfluencies, restarts, repairs, and mistakes. The transcript is, therefore, not a representation of spoken Danish but rather information content.

In the parliament hall, one speaker at a time addresses members of the parliament. Monologues may include rebuttals or other comments to statements in previous monologues. While speakers can read aloud from a prepared statement or speak extemporaneously, we expect no difference to be apparent in the data because of the post-editing.

The Folketinget section covers parliament hall sessions between 2009 and 2019. It contains discussions on a wide range of topics, issues, and named entities relevant to Danish society.

\subsection{Retsinformation}

The site \href{https://www.retsinformation.dk}{retsinformation.dk} provides access to Danish laws and regulations and documents from the Danish parliament (\ft).
The text is provided by \ft, ministries, the ombudsman of \ft, and Rigsrevisionen.
The legislative texts in this section include a variety of features: Uppercase text, redaction where names and addresses are left out, itemized text with chapter and section numbering, headlines, words with intra-letter spacing.

\subsection{Spontaneous speech}

The conversational corpus included originates from interdisciplinary research conducted within the Interacting Minds Center,\footnote{\url{http://interactingminds.au.dk}} and the Puzzle of Danish project\footnote{\url{https://projects.au.dk/the-puzzle-of-danish/}} at Aarhus University. 
Transcribed Danish speech is generally a rare kind of data, and spontaneous speech especially so; these manually transcribed conversations thus form a valuable resource.
Spontaneous and pseudo-spontaneous conversations come from various contexts, e.g., getting to know each other, solving a puzzle together, or making joint decisions. The participants have agreed on releasing anonymized transcripts of their conversations. All conversations involve two speakers, sometimes conversing face-to-face, sometimes via a chat tool.
Speech is transcribed post-hoc by native speakers.
Studies published relying on this data include~\newcite{fusaroli2012coming}, \newcite{dideriksen2019contextualizing}, and \newcite{tylen2016social}.

\subsection{Danish Wikipedia}
This section comprises a dump of Danish Wikipedia\footnote{\url{https://dumps.wikimedia.org/dawiki/}}, stripped of Wikipedia-specific markup. The content is collaboratively written by a broad range of authors and covers many specific articles that often do not exist in other languages. Most content has been roughly checked for syntactic and orthographic canonicity by editors of the Danish Wikipedia and is a rich source of region-specific named entities, often situated in full, fluent sentences.
The content is reproduced verbatim in accordance with the GNU Free Documentation License.

\subsection{Europarl}
The Europarl Parallel Corpus~\cite{koehn2005europarl} contains proceedings of the European Parliament in 21 European languages that were automatically extracted and aligned. We include the Danish part of the Europarl corpus and perform no pre-processing other than file format conversions.

\subsection{OpenSubtitles}
OpenSubtitles\footnote{\url{https://www.opensubtitles.org}} is a website where a community writes and shares subtitles for mostly big-budget movies. We extract the Danish subtitles from the OpenSubtitles section of OPUS~\cite{lison2016opensubtitles2016}. We clean the corpus to fix issues such as the capital letter I instead of the lower case letter L. We remove files that do not contain any characters specific to Danish (i.e., any of the letters \emph{\aa}, \emph{\ae}, or \emph{\o}).

\subsection{Religious text}
This section contains a Danish translation of the Bible from the Massively Parallel Bible corpus~\cite{Christodouloupoulos2015} without any  pre-processing other than file format conversion. We continue to look for other sources of religious textual content to improve the coverage and significance of this section.

\subsection{Danish Twitter}
Social media content is rich in unedited text, allowing for a very broad range of expressions. We know that social media users typically vary their language use to afford some representation for what would typically be communicated non-verbally, and while there are corpora for this for e.g. English, there are very few published corpora containing Danish social media text (e.g., ~\cite{hovy2015user,lillie2019joint}).
This section contains two datasets of Danish tweets as dehydrated content, and includes a script for rebuilding this part of the corpus, thus permitting GDPR-compliant redistribution. The first dataset contains approximately 29\,000 tweets in Danish from the \#dkpol hashtag collected during the national parliamentary elections of 2019.
The second dataset, consisting of approximately 1.6 million Danish tweets collected between April-June 2020, is not constrained by topic as tweets were collected using the 250 highest frequency Danish words.  

\subsection{DanAvis20}

Corpus DanAvis20 consists of articles from various national Danish (daily) newspapers, including Aktuelt, Berlingske Tidende, Dagen, and Weekendavisen. The articles were published during 1999-2003. All texts included have been cleared for distribution under the CC0 license (cf. Section~\ref{sec:licensing}). As part of the clearing agreement, the papers were slightly edited by limiting all text quotes to 200 words (at most), picking sentences from longer papers at random. Sentences were mildly scrambled (DanAvis20 has no instances left of 4 adjacent sentences). Proper names were pseudonymized (except ``Denmark'', ``K{\o}benhavn'', ``USA'', and a few others). Infrequent content words (10ppm or less) were replaced in situ by ``statistical cognates'', i.e., words of similar frequency and equivalent morpho-syntactic form (e.g., replacing ``Der er sardiner i k{\o}leskabet.'' with ``Der er skilsmissesager i forsikringsselskabet.'' while keeping ``Ministeren rejser hjem igen''). As overall statistical and lexical properties of DanAvis20 are thus kept invariant, the corpus still provides good material for most NLP training purposes.

\subsection{The \emph{Bornholmsk Ordbog} Dictionary Project}
Fictional texts of various kinds written in Bornholmsk, the dialect
spoken on the Danish island of Bornholm,\footnote{The language code for Bornholmsk
    under IETF BCP-47 is da-bornholm.} have been digitized (OCR'ed and proofread)
by volunteers working within the recently resumed \emph{Bornholmsk Ordbog}
dictionary project \cite{kjeldsen19}.  Most of the material included is written by Otto J. Lund in the
period 1930-48 (novels, short stories, and poems). The Bornholmsk subcorpus, which in its
present state amounts to circa 400~K words, also includes folk
stories published by J. P. Kuhre in 1938, and by K. M. Kofoed in 1935, fictional letters by various authors published in the 1930s, as well
as poems by Alfred Jensen published in 1948 and various other texts from the same period. The non-standardized orthography varies considerably from source to source. The Bornholmsk part of the Danish Gigaword is a significantly extended dataset, well beyond that studied in earlier NLP work on the dialect~\cite{derczynski2019bornholmsk}.

\section{File format}
\label{sec:app:file_format}

The philosophy is to present data as plaintext, UTF8, one file per document. Accompanying metadata gives information about (for example) the author, the time or location of the document's creation, an API hook for re-retrieval of the document, among others.

\subsection{Corpus Sections}

As the corpus many sections, per section, we do the following:

\begin{itemize}
    \setlength\itemsep{0em}
    \item
    Give each corpus section a directory with an agreed name.
    \item
    Keep all plaintext as one file per document.
    \item
    Use a section prefix, underscore, and document identifier as the
    filename, e.g.,~``tv2r\_01672''.
    \item
    Do not use file extensions for the text files.
    \item
    Maintain a one-record-per-line JSONL file in the directory, with the
    same name as the section, and with ``jsonl'' suffix, e.g.,
    ``tv2r.jsonl''. The content of this file should follow the JSONL
    format, see http://jsonlines.org.
    \item
    Each document's metadata is placed as a single JSON record in the
    JSONL metadata file, with a key ``doc\_id'' matching the filename it
    describes. Separate entries by line breaks (i.e., one JSON object per
    line).
    \item
    A \texttt{LICENSE} file should be included in each section, stating
    the license under which the section is distributed. CC and public
    domain only! Preferably CC0 or CC-BY; CC-NC if we have to. No copyleft
    licenses - they restrict the use of the data too much, which we are
    trying to avoid.
\end{itemize}

Here are the fields for the standoff JSONL metadata file entries:

\begin{itemize}
    \setlength\itemsep{0em}
    \item
    \texttt{doc\_id}: a string containing the document ID, which is also
    its filename. Begin with the section prefix, followed by an
    underscore. \texttt{String}. \textbf{Required}.
    \item
    \texttt{date\_published}: the publication date of the source document,
    including the timezone. If only the year is available, use
    \texttt{year\_published} instead. In the Python \texttt{strftime()}
    format, use \texttt{"\%c\ \%z"}. \texttt{String}. \emph{Preferred}.
    \item
    \texttt{uri}: the URI from which the document originated; can be an
    API endpoint that links directly to the data. \texttt{String,\ URI}.
    \emph{Preferred}.
    \item
    \texttt{year\_published}: the year CE that the source document was
    published. \texttt{Integer}. Use only as an alternative to
    \texttt{date\_published}. \emph{Optional}.
    \item
    \texttt{date\_collected}: the date at which the source document / API
    result collection, including the timezone. In the Python strftime()
    format, use \texttt{"\%c\ \%z"}. \texttt{String}. \emph{Optional}.
    \item
    \texttt{date\_built}: the date this document was included in the
    current version of the dataset, including the timezone. In the Python
    strftime() format, use \texttt{"\%c\ \%z"}. \texttt{String}.
    \emph{Optional}.
    \item
    \texttt{location\_name}: the name of the location of the document's
    origin. \texttt{String}. \emph{Optional}.
    \item
    \texttt{location\_latlong}: latitude and longitude of the document's
    origin. \texttt{List\ of\ two\ floats}. \emph{Optional}.
\end{itemize}

\subsection{Speech transcripts}

To represent speakers in the text files, prefix each turn with ``TALER
1:'' (substituting whatever ID is appropriate). Note: there is no space
before the colon; use one space after the colon. It is also OK to
include the speaker's name directly if this is publicly known, e.g.,
``Thomas Helmig:''.

For multi-speaker corpus sections, an optional \texttt{talere.jsonl}
file can be included in the section, containing one JSON dictionary keyed by speaker ID. Speaker IDs should be consistent through all
documents in a section. Speaker IDs need only be unique to speakers in a
section, not universally.

\end{document}